\documentclass{article}

\usepackage[numbers]{natbib}

\usepackage[preprint]{neurips_2022}

\usepackage[utf8]{inputenc} 
\usepackage[T1]{fontenc}    
\usepackage{hyperref}       
\usepackage{url}            
\usepackage{booktabs}       
\usepackage{amsfonts}       
\usepackage{nicefrac}       
\usepackage{microtype}      
\usepackage{xcolor}         

\usepackage{microtype}
\usepackage{graphicx}
\usepackage{booktabs} 
\usepackage{lipsum}
\usepackage{subfig}
\usepackage{multirow}

\usepackage{algorithm}
\usepackage{algpseudocode}

\usepackage{caption}
\usepackage{multirow}
\usepackage{graphicx}
\usepackage[normalem]{ulem}
\useunder{\uline}{\ul}{}

\usepackage{caption}
\usepackage{hyperref}

\usepackage{algorithm}
\usepackage{algpseudocode}
\algblock{Input}{EndInput}
\algnotext{EndInput}
\algblock{Output}{EndOutput}
\algnotext{EndOutput}
\newcommand{\Desc}[2]{\State \makebox[10em][l]{#1}#2}

\usepackage{amsmath}
\usepackage{amssymb}
\usepackage{mathtools}
\usepackage{amsthm}
\usepackage{listings}
\usepackage[capitalize,noabbrev]{cleveref}
\crefname{section}{Sec.}{Secs.}
\Crefname{section}{Section}{Sections}
\Crefname{table}{Table}{Tables}
\crefname{table}{Tab.}{Tabs.}

\usepackage[inline]{enumitem}

\newcommand{\gelu}{ \text{GeLU}}

\newtheorem{remark}{Remark}

\begin{document}


\title{Differentiable Outlier Detection Enable Robust Deep Multimodal Analysis}

%

\author{%
  Zhu Wang \qquad Sourav Medya \qquad Sathya N. Ravi\\
  Department of Computer Science, University of Illinois at Chicago\\
  \texttt{\{zwang260,medya,sathya\}@uic.edu} \\
}

\maketitle

\begin{abstract}
Often, deep network models are purely inductive during training and while performing inference on unseen data. Thus, when such models are used for predictions, it is well known that they often fail to capture the semantic information and implicit dependencies that exist among objects (or concepts) on a population level. Moreover, it is still unclear how domain or prior modal knowledge can be specified in a backpropagation friendly manner, especially in large-scale and noisy settings. In this work, we propose an end-to-end vision and language model incorporating explicit knowledge graphs. We also introduce an interactive out-of-distribution (OOD) layer using implicit network operator. The layer is used to filter noise that is brought by external knowledge base. In practice, we apply our model on several vision and language downstream tasks including visual question answering, visual reasoning, and image-text retrieval on different datasets. Our experiments show that it is possible to design models that perform similarly to state-of-art results but with significantly fewer samples and training time.



\end{abstract}

\section{Introduction}
Recent work shows that utilizing implicit layers  in deep network context can be significantly  beneficial for various Machine Learning (ML) tasks such as hyperparameter optimization, meta learning, and solving inverse problems in image processing settings \cite{blondel2021efficient, gilton2021deep,huang2021textrm}. In fact, if a desired input-output requirement {\em within} a ML pipeline can be formulated as a convex optimization problem, then we can use existing off-the-shelf CvxPy layers for guaranteed subgradients \cite{agrawal2019differentiable}. For discrete requirements, we can simply use the convex relaxation for backpropagation purposes. However, the existing implementations may require more memory for additional slack variables or time to handle the Jacobian, both of which are impractical in large-scale settings.  More recently, \cite{fung2022jfb} argued  that it may be possible to backpropagate through implicit layers as long as we can write it as a  Network Operator $\mathcal{N}(\cdot)$ defined as a sequential application of an operator $F(\cdot)$ that is guaranteed to  converge to a (fixed) point in some $T<\infty$ iterations. Formally, we consider network operators $\mathcal{N}(x)$ that can be written as,
\begin{align}
    \mathcal{N} (x;\Theta) = \underbrace{F\circ F\circ \cdots\circ F}_{ T \text{ times}} (x;\Theta),\label{eq:ntwoper}
\end{align}
where $\Theta$ corresponds to parameters of a learning model such as the deep network. Network {\em Operators} in \eqref{eq:ntwoper} are convenient as it turns out that we can backpropagate through such $\mathcal{N}$ with neither its full sequence (or trajectory) nor solving an inverse problem with its Jacobian for gradients \cite{fung2022jfb}. Now, to see the utility of $\mathcal{N}(x;\Theta)$ in  ML settings,  we consider the task of Trajectory prediction in Vision settings. Here, by using $\mathcal{N}$ (with an appropriate $F$) as a  clustering map on training data, we can easily define performance improving loss functions. \cite{sun2021three}.    
\begin{figure*}[!t]
\includegraphics[width=0.9\textwidth]{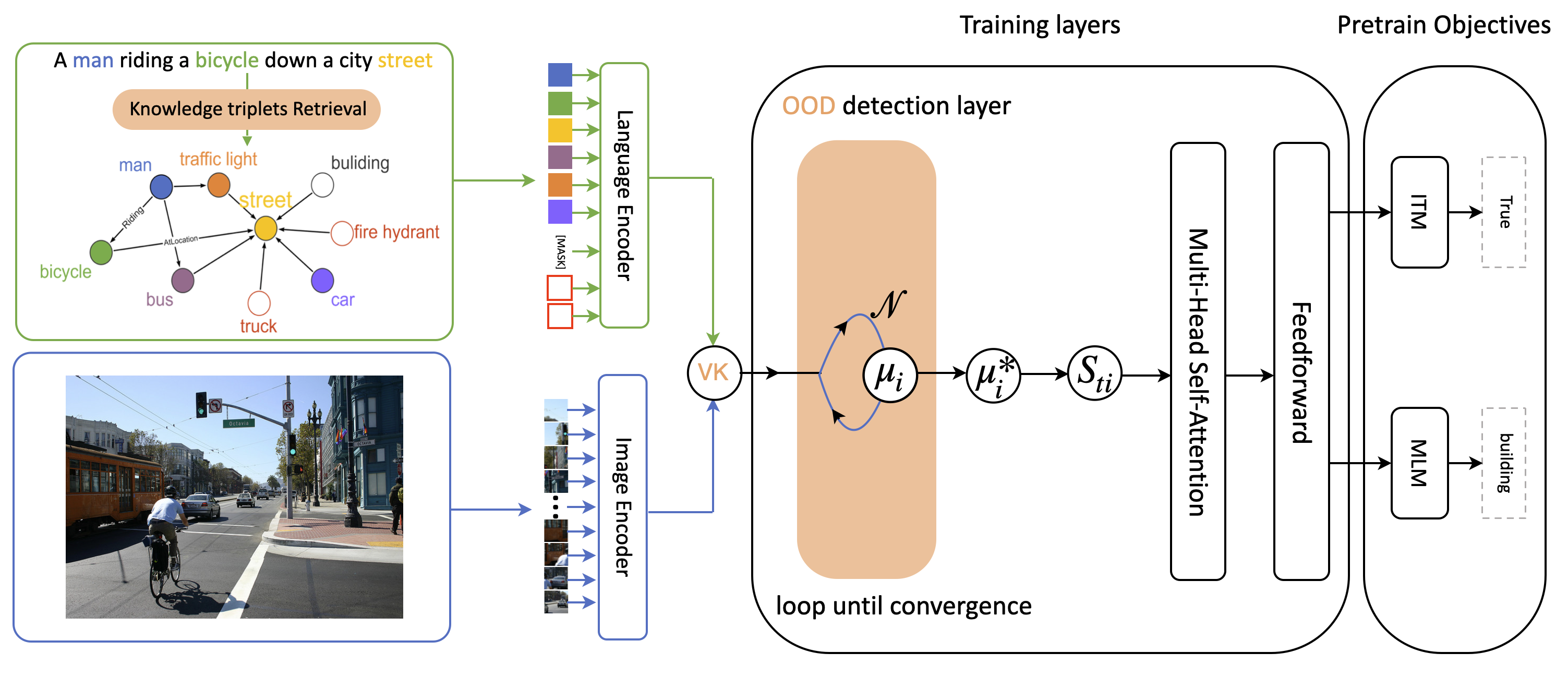}
\centering
\caption{\label{fig:model}%
Model architecture of our proposed method. Inputs are an image and a caption. Image is segmented to flatten patches, and the caption is parsed and integrated with knowledge graph. Image-triplet pairs are fed into an OOD detection layer. This layer detects noise concepts before passing features to the cross-attention multimodal transformer encoder. Our pretrain objectives contain ITM and MLM which represent image text matching, and masked language modeling respectively.}
\end{figure*}
\setlength{\dbltextfloatsep}{5pt}

In multimodal analysis, a model such as a deep network with parameters $\Theta$ is trained to align multiple modalities such as visual and language embedding spaces,  \cite{radford2021learning,ngiam2011multimodal}. After learning  $\Theta$, one can use the deep network to predict the alignment of the modalities. For large-scale multimodal tasks, \cite{zhang2020multimodal, kim2021vilt} provide guidelines for deep network architectures, and state of the art strategies for fusing modalities. 

In this paper, we consider the problem of designing Network Operator $\mathcal{N}$ for  incorporating external knowledge in large-scale  Multimodal pipelines. Intuitively, explicit knowledge corresponds to extracting the relevant part of the idea while ignoring others which is often done by considering the fidelity with multiple other modalities of the same input data. Moreover, the explicit information brings the reasoning ability to the multimodal analysis \cite{gui2021kat}. Indeed, there have been many solution schemes to overcome technical challenges when features are from multiple modalities  simultaneously \cite{chen2020uniter, li2021align}.  However, these strategies are impractical when used in-tandem with features derived from external knowledge graphs that are {\em prone} to noise. For example, the fusion models in \cite{gan2020large, zhang2021vinvl} often fail to filter noise pairs when pre-training on large-scale datasets, leading to slow convergence. Specifically, there may be objects or concepts that are anomalies in image-text pairs. For example, given an object ``street'' and relation ``locatedAt'', noisy (since absent in the input image) objects such as ``fire hydrant''  are returned with high confidence by ConceptNet \cite{speer2017conceptnet}, undesirable for training purposes.

 Our key technical idea to handle noise from external knowledge as described above is to score the features derived from KGs using an out-of-distribution (OOD) detection layer within the pipeline. We show how such a layer can be written as a network operator $\mathcal{N}$ in equation \eqref{eq:ntwoper} for efficient backpropagation. By deriving features corresponding to the relevant external knowledge bases using a large pre-trained language model \cite{liu2019roberta,brown2020language}, we show how to accelerate training multimodal models seamlessly. To do so, we approximate the density of in-distribution features using the concepts in the caption corpus using recently developed out-of-distribution scores \cite{morteza2022provable} for all concepts. Thus, in this work, we provide an end-to-end training framework with knowledge graphs and multimodal models in a differentiable manner, as illustrated in Figure \ref{fig:model}. 

To showcase the effectiveness of our proposed OOD detection layer based on $\mathcal{N}$, we perform several experiments using our pre-trained vision-language models. To demonstrate the practical benefits of KGs in large-scale multimodal pipelines, we perform extensive fine-tuning experiments on several tasks such as visual question answering, and image text retrieval. {\bf We will now briefly summarize our contributions here:}

\begin{itemize}
\item First, we show integrating explicit knowledge benefits multimodal fusion models. Our pre-train model learns from visual and textual representations and incorporate implicit and explicit knowledge seamlessly.

\item Second, our proposed OOD detection implicit layer can perform outlier detection task with an efficient backpropagation manner in practical settings.

\item  Third, our model outperforms six baselines on three downstream tasks on several datasets. Moreover, we provide several ablation studies and an interactive user study which shows that users is able to sustain their desired in-distribution like inputs or features. 



\end{itemize}

\section{Related Work}
\textbf{Vision-and-language transformer.} Most recent vision-and-language models have shown promising performance in self-supervision learning as attention-like mechanism for multimodal tasks. These models are pre-trained on large scale image-text pairs and then finetuning on downstream VL tasks. UNITER  \cite{chen2020uniter}, VinVL \cite{zhang2021vinvl}, and PixelBERT \cite{huang2020pixel} introduce the cross-modal architecture to learn joint multimodal representations of visual and textual contents. ViLT \cite{kim2021vilt}, ALBEF \cite{li2021align} and FLAVA \cite{singh2022flava} directly use image patch features extracting from backbone and align with text features passing into a multimodal encoder. However, these apply the strategies of scaling up the model by feeding more image-text pairs or more layers. While a large model with more parameters can be effective, it is inefficient for image and text retrieval tasks on large scale data and the latent semantic information is still left to be explored.   

\textbf{Knowledge-based Multimodal analysis.} In the area of vision and language understanding, knowledge representation learning has been capitalized on image and text or sentence matching \cite{huang2018learning, garderes2020conceptbert, wang2020consensus}. These works aligned extracted consensus or semantic concepts with visual concepts, but they applied statistical method to construct concepts vocabulary. Recently, KRISP \cite{marino2021krisp}, MAVEx \cite{wu2022multi}, UnifER \cite{guo2022unified} and KAT \cite{gui2021kat} proposed knowledge based VQA to integrate explicit knowledge. However, these mainly focus on VQA settings, the effectiveness of incorporating external knowledge in vision-language pre-training models still remains to be explored. 


\textbf{Out-of-distribution (OOD) detection.} 
We discuss the out-of-distribution detection methods in training deep neural networks. GEM \cite{morteza2022provable} has shown a provable OOD detection method to estimate the distributions. They demonstrate the GEM score comparing with multiple prior OOD detection to provide the theoretical and empirical guarantees. It is beneficial to use GEM score to recognize irrelevant distributions of concepts during training. However, previous works focus on the detection performance and it is unclear how to utilize trustworthy OOD detection methods for improving pre-train model performance in practice. 

\section{Knowledge-based OOD detection Multimodal Pipeline}
To integrate explicit knowledge, we propose a novel architecture called \textbf{VK-OOD} for multimodal analysis by fusing \textbf{V}ision and \textbf{K}nowledge features. By using an {\bf O}ut {\bf O}f {\bf D}istribution (OOD) detector as the network operator $\mathcal{N}$, we detect concepts that can potentially lead to slow convergence of upstream and/or downstream layers during training. The model architecture is constructed by vision and knowledge with texts end-to-end transformer encoders and shown in Fig \ref{fig:model}. VK-OOD aims at learning the joint representations of images and texts integrating commonsense knowledge while filtering out noisy concepts. Our method utilizes external background knowledge graph encoding commonsense facts of the objects and relations about the objects in the input images and captions. Our OOD network operator can can be adapted to various downstream tasks. 


\textbf{Architecture.} Given an image $I$ with a caption (question) $S$ in the form of a sentence, our pipeline consists of the following steps, implemented as differentiable modules in our VK-OOD architecture:
\begin{enumerate}
\item Extract patches from Images, and transform to  informative features using standard vision models,
\item Retrieve external knowledge triplets using Knowledge Graphs, and transform to language features,
\item While in the pipeline, we approximate density of in-distribution features to filter each image-triplet pair (including ones derived from external knowledge) into an OOD detection layer to filter outliers,
\item Finally, we learn vision and language representations by a multi-modal cross-attention transformer with multiple training objectives.
\end{enumerate} 

\textbf{Obtaining Image Features Using Encoder Module. } We segment  images into $P$ number of patches and extract $V$ as image features by vision encoder. We denote the features extracted from the patches by $v_m,m=1,...,P$.

\textbf{Knowledge Triplets Retrieval Module. } We utilize semantic information from the captions and derive triplets from external knowledge graph. We provide more details of retrieval process in Appendix \ref{ap:encoder}. Then, similar to vision features, we denote the final encoded features by $l_{ti},ti=1,....T^{cn}$.


\subsection{Defining OOD detection layer as a Fixed Point Operator} Given Vision features $v_m$, and Language features $l_{ti},$ the goal of OOD detectionn layer in our VK-OOD model is to compute a score $s_{ood}^{ti}\in\mathbb{R}_+$ for each language feature $l_{ti}$. To do so during training, we will now explain how to characterize the  density or distributions  of the {\bf in-distribution} (ID) triplet features for OOD purposes. 

{\bf Finding Fixed Points for Forward Pass.} Finite mixtures with $k$ components are conceptually simple. Moreover, theoretically they can produce accurate
approximations to most density functions \cite{mclachlan2019finite}. So, we approximate the density of ID features using a Gaussian Mixture Model (GMM) and use $\mu_i^*,\Lambda_i^*=1,\dots,k$ to denote the (unknown) optimal means and covariance matrices of $k$ components. 

For GMMs, we solve for $\mu^*$ is by using the standard Expectation Maximization (EM) algorithm \cite{murphy2022probabilistic}. Our main observation is that the update rule used in an EM algorithm on the current iterate $\mu_i$ can be written as a fixed point iteration as follows:\begin{align}
    \mu_i\leftarrow \frac{\sum_{ti=1}^{T^{cn}}\exp(-w(\mu^{ti}_i))l_{ti}}{\sum_{ti=1}^{T^{cn}}\exp(-w(\mu_i^{ti}))}\label{eq:gmmem}
\end{align}
where the weight of current iterate $\mu_i$ on $l_{ti}$ denoted by $w(\mu_i^{ti}):=\sum_{i=1}^{k}\|\Lambda_i^{-0.5}(l_{ti}-\mu_i)\|_2^2$. { We update  $\Lambda_i$ similarly using $(l_{ti}-\mu_i)(l_{ti}-\mu_i)^T\in\mathbb{R}^d$ in the numerator of fixed point operation in Equation \eqref{eq:gmmem}.} This observation implies that we can get an approximation of in-distribution density. Now we explain why EM algorithm is our preferred choice from the theoretical standpoint to construct the network operator $\mathcal{N}$ for OOD detection.

{\em Benefits of EM algorithm.} EM updates are provably convergent in various settings, and it takes few iterations when it is guaranteed to converge \cite{daskalakis2017ten}. This also implies that we can simply initialize $\mu_i$ randomly and perform few more iterations. Moreover, since each iteration in GMM \eqref{eq:gmmem} is differentiable, we can easily backpropagate through few iterations of EM algorithm. However, since the denominator contains terms that are also unknown parameters, such update schemes may be numerically unstable. In such cases, we can simply use recently proposed gradient based  EM algorithms as the network operator, for fine tuning deep networks in large scale settings, see \cite{segol2021improved} for convergence analysis. 

\begin{algorithm}[!tp] 
\caption{Fixed Point Network Operator based Out-of-Distribution Detection  Layer for Language Features $l_{ti}$}\label{alg:cap}
\begin{algorithmic}
\Input
  \Desc{$l_{ti}\in\mathbb{R}^d~ \forall ~ti\in[T^{cn}]$} \Comment {Language Features}
  \Desc{$\mu_i\in\mathbb{R}^d~\forall~i\in[k]$}\Comment{Initial GMM Means}
  \Desc{$\Lambda^0_i\in\mathbb{R}^{d\times d}~\forall~i\in[k]$}\Comment{Initial GMM Covariances}
  \Desc{$\lambda_1> 0$ in Equation \eqref{eq:threshold}}\Comment{OOD threshold parameter }
  \Desc{$T\in\mathbb{N}$}\Comment{Maximum Iterations}
  \EndInput
  \Output
  \Desc{$G_{ti}\in\{0,1\}$}\Comment{Featurewise OOD Indicator}
  \Desc{$\nabla_{l_{ti}} G_{ti}(l_{ti},\mu_i^*)$}\Comment{Featurewise Gradients}
  \EndOutput
  \State {\em -- Begin Forward Pass --}
\While{$t \leq T$}
    Update $\mu_i$ using Equation  \eqref{eq:gmmem}
\EndWhile
\State {Set $\mu_i^*$ to be the last iterate of $\mu_i$}
\State {\bf OOD Detection.} Compute $G_{ti}$ in Equation \eqref{eq:threshold}
\State {\em -- Begin Backward Pass --}
\State {{\bf Jacobian Free Backpropagation.} Output gradient $\nabla_{l_{ti}} G_{ti}(l_{ti},\mu_i^*)$ by computing derivative of composition of log-sum-exp  and ReLU functions in Equations \eqref{eq:ood_score} and \eqref{eq:threshold} using Chain rule.}
\end{algorithmic}
\end{algorithm}

\setlength{\textfloatsep}{5pt}

{\bf GEM score for Efficient Backpropagation.} So far, our forward pass computes the optimal means $\mu^*$ and covariances $\Lambda^*$ of the GMM approximation of the ID feature density function using a fixed point or network operator.  Having found $\mu_i^*,\Lambda_i^*$, we can view the initial part of our OOD network operator $\mathcal{N}_{ood}$ as the EM algorithm that outputs the optimal parameters of ID feature density. But EM viewed as an operator from $\mathbb{R}^d\to \mathbb{R}^d$ (mapping $l_{ti}\in\mathbb{R}^d$ to $\mu_i^*\in\mathbb{R}^d$) makes backpropagation tricky since the Jacobian of such a map will be a $\mathbb{R}^{d\times d}$ matrix, practically infeasible for training purposes even when $d\approx 100$ is not very large.   The final ingredient we need for samplewise, and memory efficient forward pass is to be able to compute a score for each $l_{ti}$  language feature, for which we rely on the recent statistical developments in OOD detection, that have already been tested on some simple classification applications. We use the recently introduced GEM score in \cite{morteza2022provable} to filter anomalous triplets to obtain memory efficient gradients, shown in Figure \ref{fig:model}.   Given an derived feature from a knowledge triple $ti \in T^{cn} $, its GEM score is defined using a log-sum-exp energy function as,
\begin{align}\label{eq:ood_score}
    s_{ti} = \log \sum_{i=1}^k \exp{(-\frac{1}{2}(l_{ti} - \mu_i^*)^T \Lambda_i ^{-1}(l_{ti}-\mu_i^*))}
 \end{align}
where $l_{ti}\in\mathbb{R}^d$ is the language feature of external triplets, $\mu_i^* \in \mathbb R ^d$ is the output of the network operator $\mathcal{N}_{ood}$ (as in Eq \eqref{eq:ntwoper}) of ID triplets $\{t_1,...t_k\}\in T$ and $\Lambda^* \in \mathbb R^{d \times d}$ is the covariance matrix of ID triplets. The corresponding OOD detection $G_{ti}\in\{0,1\}$ using Eq\ref{eq:ood_score} is given by simple thresholding,
\begin{align}\label{eq:threshold}
G_{ti} =1, & \quad \text{if } s_{ti} \geq \lambda_1,
\end{align}
where $\lambda_1$ denotes a threshold parameter, when $G_{ti} = 1$, $l_{ti}$ will be concatenate to ID triplets, otherwise to negative image-text pairs. $\mu_i$ is updated during the training process, so $s_{ti}$ is a trainable parameter to better capture uncertainty.  { We can define a OOD procedure for $v_m$'s similarly.}

\textbf{Interaction via Outlier Detection.} For applications that require a high number of patches $P$ (or concepts $cn$),  the likelihood that one of the patch features or text features to be an outlier also increases dramatically. In high dimensional settings, this can increase the training time taken by first order methods significantly, especially when minibatches are used to compute gradients \cite{geiping2021stochastic}. Alternatively, when features $l_n$ (or $v_m$) are computationally easy to extract, say using a GAN, it is reasonable to expect that a certain fraction of the $l_n$ or $v_m$ are outliers, and should not be used for backpropagation purposes. In a more optimistic scenario, we may want to customize our predictions, and handle ``on-the-fly'' integration of explicit knowledge.  In our framework, this corresponds to treating  $\mu_i$ in \eqref{eq:ood_score} as trainable parameters. We can update the initialization $\mu_i$ without storing the trajectory, or forming the full Jacobian which can be expensive, as in our Algorithm \ref{alg:cap}.

\subsection{Multimodal Training using ID Concepts}  Now we will explain how to setup the overall training objective using the OOD indicators $G_{ti}$. For this, we follow standard procedures in which the features from modalities are matched. First, we compute pairwise {\bf signed} scores $s_{mti}\in(-1,+1)$ by applying elementwise nonlinearity. These scores $s_{mti}$ are  then used to obtain    {\bf unsigned} weights $\alpha_{mti}\in(0,1)$ by  applying a SoftMax operation  with an inverse temperature  $\lambda_2>1$ (see \cite{chorowski2015attention} for more details).  That is for each image-language feature pair $v_m,l_{ti}$, we set,
\begin{align}
\begin{split}
s_{mti} = \gelu\left(\frac{v_m^Tl_{ti}}{\|v_m\|\|l_{ti}\|}\right),\alpha_{mti}=\textbf{softmax}(s_{mti})\label{sim_score_1}
\end{split}
\end{align}
where  $\gelu(x) :=x\mathbb{P}\left(\mathcal{N}(0,1)\leq x\right)$,  here $\mathcal{N}(0,1)$ is the standard normal distribution, and $\textbf{softmax}$ returns a distribution over $ti$ for each image patch $m$, that is, each $ \alpha_m\in\mathbb{R}^{T^{cn}}$ is a nonnegative vector, and sums to 1. In our implementation, we simply  reweigh encoded language features using  a vision information from image encoders. Specifically, we  reweigh the text features $l_{ti}$ by summing over all the unsigned weights to obtain 
\begin{align}
a_{ti}^t = \sum_{m=1}^P {\alpha_{mti}l_{ti}} = l_{ti}  \sum_{m=1}^P {\alpha_{mti}},\nonumber
\end{align}
Finally, we calculate the similarity between  these reweighed language features $a_{ti}^t$, and vision features $v_m$ as,
\begin{align}
R(v_m,a_{ti}^t) = \frac{v_m^Ta_{ti}^t}{\|v_m\|\|a_{ti}^t\|}, m\in[P].
\label{eq:sim_score_4}
\end{align}
Intuitively, for a test or unseen image-caption sample, a large value of $R$ in equation \eqref{eq:sim_score_4} indicates that patch $m$ and concept $ti$ are likely to occur together for a fixed set of learnable parameters. With training data given as $(v_m,l_{ti},y_{mti})$, we can simply use  such similarity based loss functions even under the presence of outliers with the help of $G_{ti}$ computed using our OOD Layer \ref{alg:cap}.  Hence, the loss function we use for concepts $ti$  in our VK-OOD architecture can be written as,
\begin{align}\label{eq:lossood}
    \mathcal{L}_{\text{ood}} = \mathbb{E}_{(V,L_{KG}) } G_{ti}  \mathcal{H} (y_{mti},R(v_m,l_{ti})),
\end{align}
where $L_{KG}$ is the distribution of language features obtained from external knowledge, $\mathcal{H}$ denotes the usual cross-entropy function, $y_{mti}$ corresponds to the matching label of image and extracted knowledge triplets. 

{\bf Overall Training Objective.} We use our external knowledge based loss in equation \eqref{eq:lossood} with two standard matching based loss functions commonly used in Multimodal training pipelines. First, we use {\bf Image Text Matching (ITM)} loss $\mathcal{L}_{\text{ITM}}$ defined as,
\begin{align}\label{eq:itm}
    \mathcal{L}_{\text{ITM}} = \mathbb{E}_{(V,L) } \mathcal{H}(y_{mn},R(v_m,l_n)) +\mathcal{L}_{\text{ood}},
\end{align}
and second, we use {\bf Masked Language Model (MLM)} loss $\mathcal{L}_{\text{MLM}}$  defined as,
\begin{align}\label{eq:mlm}
    \mathcal{L}_{\text{MLM}} = \mathbb{E}_{(V,\hat{L}) } \mathcal{H}(y_{mn},R(v_m,\hat{l}_n)) +\mathcal{L}_{\text{ood}}
\end{align}
where $\hat{L}$ denotes the distribution of language features obtained from masked tokens. The total loss function $\mathcal{L}$ of our VK-OOD model we propose  is a linear combination of  the ITM and MLM loss  in equations  \eqref{eq:itm}, and \eqref{eq:mlm} is given by, 
\begin{align} 
    \mathcal{L} = \mathcal{L}_{\text{ITM}} + \lambda_3 \mathcal{L}_{\text{MLM}}
\label{loss}
\end{align}
where  $\lambda_3>0$ is a regularization parameter.
\setlength{\textfloatsep}{5pt}
\begin{remark}
Note that, $s_{ti}$ of the same triplets in different training objectives may vary on random masks. Therefore, we add $\mathcal{L}_{\text{ood}}$ in both Eq \ref{eq:itm} and Eq \ref{eq:mlm}. We also provide details of our pre-train objectives with OOD loss in Appendix \ref{ap:pre-train}.
\end{remark}

\section{Experiments on Downstream Tasks}
In this section, we introduce datasets and implementation details in pretrain and finetune settings. To evaluate our model, we conduct experiments on multiple downstream tasks, including (1) visual question answering, (2) natural language for visual reasoning, and (3) image-text retrieval.

\subsection{Implementation details} \label{sec:imp}
\textbf{Datasets.} We pre-train on three datasets, including COCO \cite{lin2014microsoft}, Visual Genome \cite{krishna2017visual}, and SBU Captions \cite{ordonez2011im2text} with total of 1M images and 6.8M image-caption pairs, as approximate 30$\%$ less than baselines. Each caption is parsed to 1 - 3 triplets and augmented with 5 external knowledge triplets. For downstream datasts, we use Flickr30k \cite{plummer2015flickr30k} and COCO for image-text retrieval, VQAv2 \cite{antol2015vqa} and OKVQA \cite{okvqa} for visual question answering and ablation studies, and NLVR2 \cite{suhr2018corpus} for visual reasoning. We resize each image to the size of $224\times224$ by center-cropping. In the merged attention module, each multimodal encoder layer consists of one multi-head self-attention block and one feedforward block, and total number of identical layers is 12. For downstream tasks, we fine-tune with base learning rate of 5e-6 on higher resolutions of image with size of $480\times480$ to obtain better performance.

\textbf{Encoder backbones.} First, we retrieve explicit knowledge triplets in pre-processing, by using ConceptNet Numberbatch\footnote{\url{https://github.com/commonsense/conceptnet-numberbatch}}. Next, we use RoBERTa \cite{liu2019roberta} and CLIP-ViT-B32 by \cite{radford2021learning} as text encoders. For the image encoder, we use CLIP-ViT-B32 \cite{radford2021learning} and Swin-Base \cite{liu2021Swin} as backbones.

\textbf{Network training.} For the pre-training, we use AdamW optimizer designed by \cite{loshchilov2018decoupled} with the base learning rate of 1e-5 for image and text encoders, and 5e-5 for multimodal module. We pre-train and fine-tune on 8 NVIDIA RTX 2080Ti GPUs. The warm-up ratio of learning rate is 10\% of the total training steps, and the learning rate was decayed linearly to 0 in the rest steps. 

\begin{table}[]
\caption{Overall performance and comparison with other models on multiple downstream tasks. ``B" denote base models. Here, we choose to compare with similar pre-train dataset size models. The bold values mean the best model in the table. Our model outperforms baseline models in all downstream tasks.} \label{tab:down2}
\begin{center}
\begin{tabular}{l|c|c|cc}
\hline
\multicolumn{1}{c|}{\multirow{3}{*}{Model}} & \multirow{3}{*}{VQAv2} & \multirow{3}{*}{NLVR2} & \multicolumn{2}{c}{COCO}                          \\ \cline{4-5} 
\multicolumn{1}{c|}{}                       &                        &                        & \multirow{2}{*}{IR R@5} & \multirow{2}{*}{TR R@5} \\
\multicolumn{1}{c|}{}                       &                        &                        &                         &                         \\ \hline
UNITER-B                                    & 72.7                   & 75.8                   & 78.5                    & 87.4                    \\
ViLT-B                                      & 70.3                   & 74.6                   & 72.0                    & 86.2                    \\
ALBEF(4M)                                   & 74.5                   & 80.5                   & 81.5                    & 91.4                    \\
VinVL-B                                     & 75.9                   & 83.1                   & 83.2                    & 92.6                    \\
PixelBERT                                   & 74.5                   & 77.2                   & 77.5                    & 87.5                    \\
FLAVA                                       & 72.8                   & 79                     & -                       & -                       \\ \hline
\textbf{VK-OOD}                             & \textbf{76.8}          & \textbf{83.9}          & \textbf{83.6}           & \textbf{93.1}           \\ \hline
\end{tabular}
\end{center}
\end{table}
\subsection{Results on Downstream Tasks} \label{sec:down}
We evaluate our VK-OOD models on common vision-language downstream tasks.  We finetune our model for 10 epochs with base learning rate of 1e-5 for all downstream tasks. In addition, we apply RandAugment \cite{cubuk2020randaugment} as augmentation strategy in finetuning steps. 

{\bf Takeaway.} In all the experiments, we establish that our VK-OOD model achieves the best performance comparing with \textbf{six SOTA} vision-language models on \textbf{three downstream tasks} with several datasets while using lower number of parameters (see Table \ref{tab:down2}).


\textbf{Visual Question Answering (VQA).} For VQA tasks, the goal is to predict answer classes as a classification problem.  We finetune our model on VQAv2 train sets and use the validation images and their questions pairs for internal validation. Then, we evaluate this task on the VQAv2 test set. The VQAv2 dataset results \footnote{\url{https://eval.ai/challenge/830/overview}} are shown in Table \ref{tab:down2}. Our model VK-OOD outperforms all the baselines on this dataset and produces 76.8\% accuracy. 



\textbf{Natural Language for Visual Reasoning.} This task is to predict whether a text description is matched to a pair of images. We evaluate our model along with the baseklines models on the NLVR2 dataset for this task. We finetune our pre-train model with the pair method that we concatenate the features of each pair (question, one image) extracting from VK-OOD and predict outputs as a binary classifier. Table \ref{tab:down2} shows the results. 
Our model VK-OOD achieves the best result in terms of accuracy with 0.8\% and 9.5\% increase than the best and worst performing baselines respectively. The performance of our model shows the ability of visual reasoning while incorporating implicit (e.g., semantic information) and explicit knowledge (e.g., knowledge graph).

\textbf{Image-text Retrieval.} The tasks include two different types: i) to retrieve images from text queries, and ii) retrieving texts from images. We evaluate our model along with the baseline models on the COCO and F30K dataset. 
Our model produces the best performance and outperforms the best and worst performing baselines by up to .5\% and 11.6\% respectively on the COCO dataset (Table \ref{tab:down2}). In other settings and on F30K dataset, the results are similar. The details are given in the Appendix (see Sec. \ref{ap:ablation})

\begin{figure}
\includegraphics[width=\textwidth]{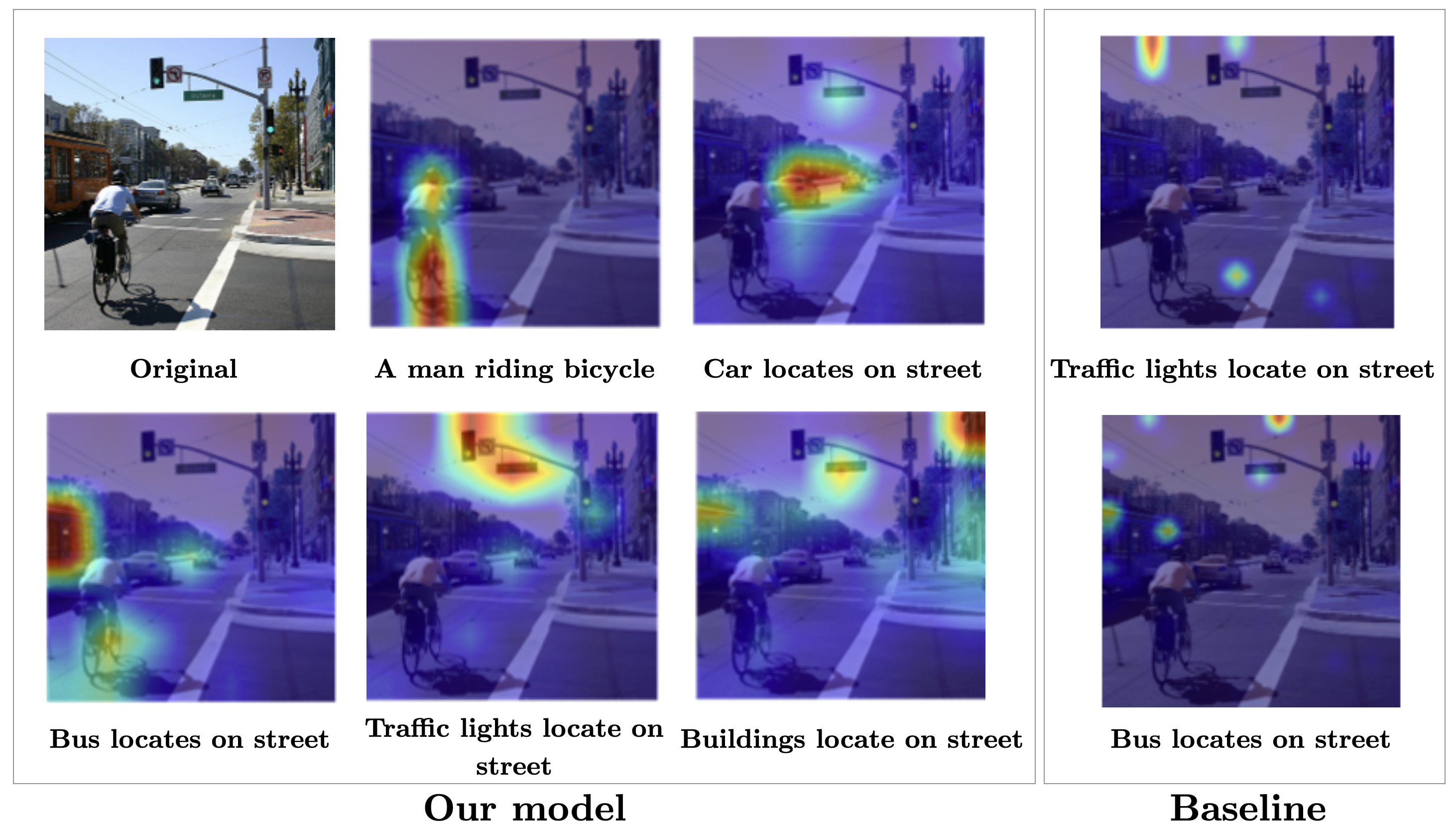}

\centering
\caption{Visualization of the attention maps of image and knowledge concept triplets alignment. The results are from our VK-OOD model. The original sample caption is ``A man riding a bicycle down a city street". We highlight areas in the example image corresponding to different knowledge triplets. Comparing with the attention maps of the baseline model, our model learns more objects and localize those objects correctly. Note that, the baseline model is trained winthout KG and OOD components.\label{fig:example1}%
}
\end{figure}

\subsection{Qualitative Analysis}
\label{sec:qualitative}

Fig. \ref{fig:example1} is an example of multimodal alignment results from our VK-OOD comparing with the model (baseline) without knowledge graph representations and OOD detection layer. We use Grad-cam \cite{selvaraju2017grad} to visualize the multimodal maps of the pre-train models on image corresponding to knowledge triplets. As shown in Fig. \ref{fig:example1}, our pre-train model has the capability to attend to the extracted knowledge concepts, such as buildings and cars. Thus, our model can detect more objects to provide the ability for answering  open questions. By contrast, the baseline model is not able to distinguish objects which are not appearing in the captions. Furthermore, we discuss more user studies on interactive OOD detections by feeding in domain knowledge with different distributions in the Appendix \ref{ap:ablation}.

\begin{figure*}
\includegraphics[width=\textwidth]{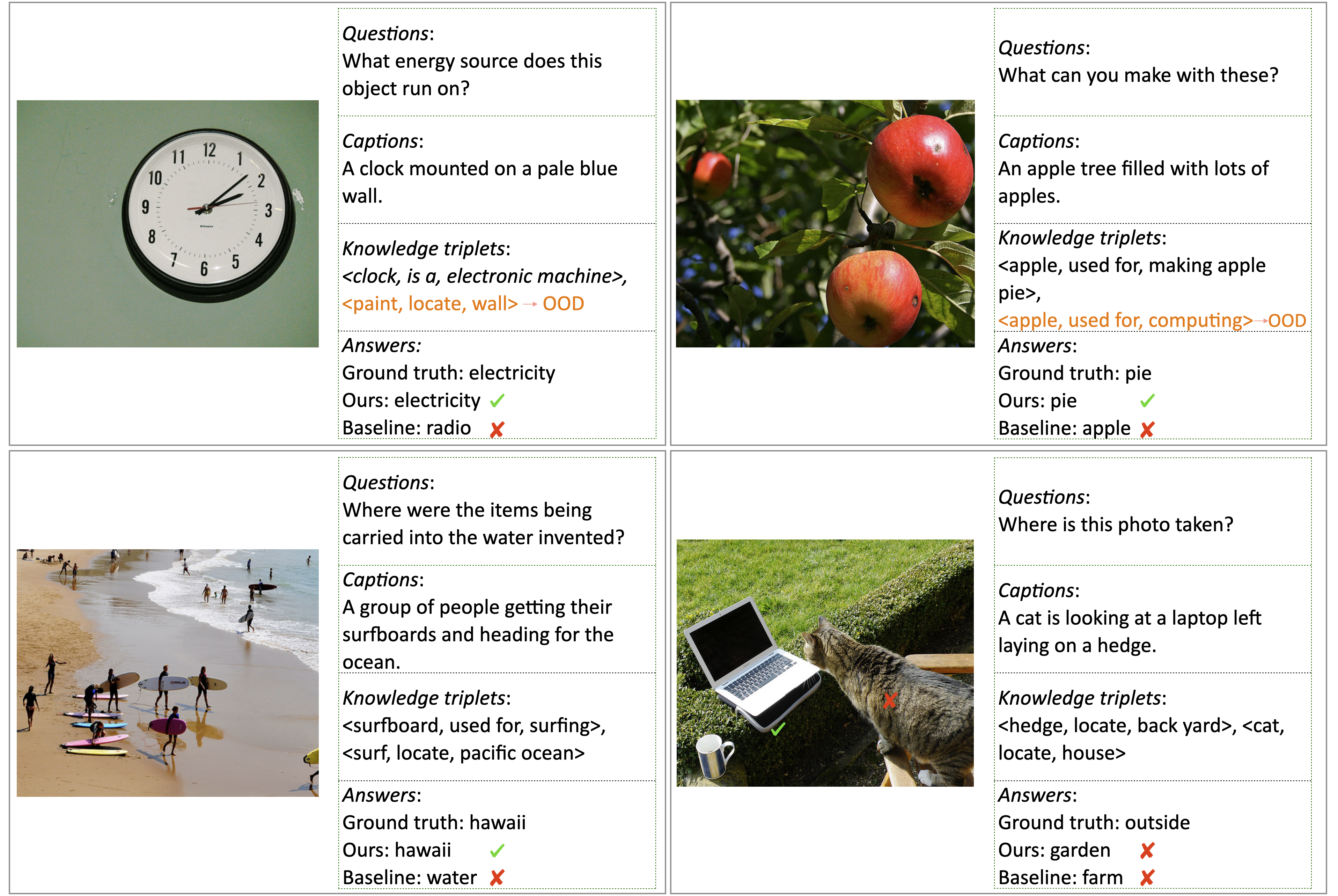}
\centering
\caption{Example case studies with OK-VAQ dataset. We show retrieved explicit knowledge triplets. The number of example triplets is 2. Our model is able to detect outliers of retrieved triplets shown in the first row examples. The predicted answers are from our proposed VK-OOD model finetuning on OKVQA dataset. Comparing with the baseline results, our model provides more correct answers. Note that, the baseline model is trained without KG and OOD components.\label{fig:example2}%
}
\end{figure*}

\begin{figure*}
\centering
\includegraphics[width=\textwidth]{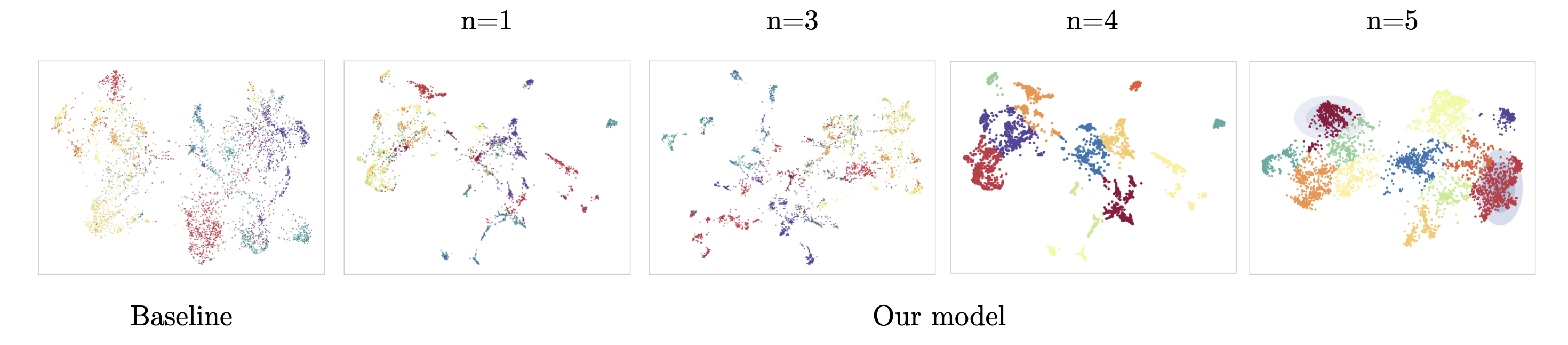}
\caption{Visualization of the multimodal feature space. $n$ denotes the number of clusters. \label{fig:feature}%
}
\end{figure*}

We demonstrate a case study with our proposed VK-OOD model on OK-VQA dataset, and visualize various results in Fig. \ref{fig:example2}. We show the extracted knowledge triplets based on the captions. For instance, \textlangle{apple, used for, making apple pie}\textrangle{} is useful to obtain correct answers comparing with the baseline model. This observation validates that explicit knowledge provides more reasoning capability than implicit knowledge. Moreover, our model detects OOD triplets by interacting with visual modality, i.e, the apple is fruit in image, thus is not used for computing. The last one is a failure case of our model, because the ground truth answer is abstract. Therefore, it brings our attention to consider more inference and reasoning abilities in multimodal analysis.


\section{Ablation Study Experiments}
We study the effectiveness of different components (knowledge triplets and OOD detection layer) from VK-OOD via several experiments. Here, we perform the following four different ablation studies: (i) the performances of different combinations of the components in our proposed model, VK-OOD (Sec. \ref{sec:ablation_vkood}), (ii) the impact of different knowledge triplets (Sec. \ref{sec:ablation_kg}) (iii) the impact of different knowledge encoders (Sec. \ref{sec:ablation_kg}) and (iv) the robustness of the OOD layer (Sec. \ref{sec:robust_ood}). Furthermore, we experiment with different backbones for image-text retrieval tasks (please see \ref{ap:ablation}).



\subsection{Ablation Study on VK-OOD Components}
\label{sec:ablation_vkood}
To compare the impact of the proposed components in our {VK-OOD} model, we consider different combinations of inclusion and exclusion of knowledge graph representations (KG) and out-of-distribution detection layer (OOD). The results are shown in Table \ref{tab:ablation1}. The results show that our model achieves the best performance when both the components are included in the model. Compared to the other settings, it produces Moreover, comparing the results on VQAV2 and OKVQA datasets, the results imply that only the external knowledge triplets (KG) can be beneficial to improve the performance especially on the visual question answering task. Furthermore, using OOD layer solely has good performance. This shows that including OOD layer in our model is helpful and able to capture the noise of multiple modalities, such as missing or mismatching modalities. 

\begin{table}[ht]
\caption{Ablation studies of different components of our proposed models. ``KG" and ``OOD" denote knowledge graph representations and out-of-distribution detection layer respectively. The setting with both components outperforms other settings on all datasets and downstream tasks.} \label{tab:ablation1}
\begin{center}
\begin{tabular}{ll|lll}
\hline
\multicolumn{2}{c|}{Method} & \multicolumn{3}{c}{Downstream tasks} \\ \hline
KG           & OOD          & VQAV2      & OKVQA      & NLVR2      \\ \hline
             &              & 73.9       & 45.5       & 80.6       \\
\checkmark           &              & 74.6       & 48.3       & 81.8       \\
             &  \checkmark            & 74.1       & 46.2       & 81.1       \\
   \checkmark          &     \checkmark         & \textbf{76.8}       & \textbf{52.4}       & \textbf{83.9}       \\ \hline
\end{tabular}
\end{center}

\end{table}


\begin{figure}
\centering

\subfloat[]{\includegraphics[width=4.55cm]{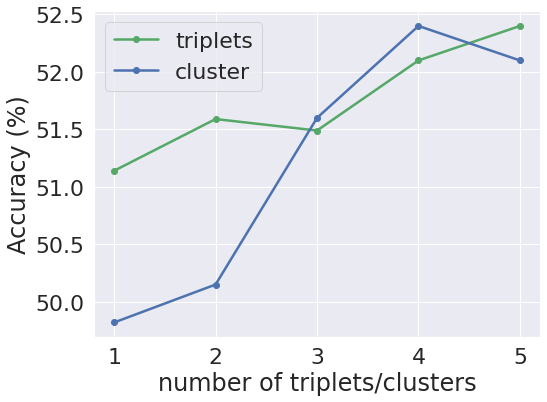}\label{fig:number}}
\subfloat[]{\includegraphics[width=5cm]{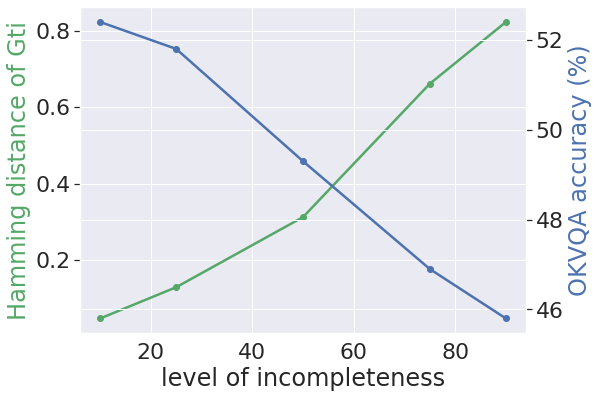}\label{fig:robustness}}

\caption{Ablation studies results on OKVQA dataset. (a) Results on accuracy with different numbers of external knowledge triplets (blue) and the number of clusters (orange) in the GMM process (see Eq. \ref{eq:gmmem}). The accuracy increases with the increase in the number of triplets. (b) Results on the robustness analysis of the OOD layer with different level of data incompleteness. The green line and blue line are the normalized Hamming distance of $G_{ti}$ (left) and VQA Accuracy (right). The green line shows the hamming distance increases with the amount of incompleteness in the data.}\label{fig:Label}
\end{figure}

\subsection{Ablation study on Knowledge Graphs}
\label{sec:ablation_kg}
We conduct multiple ablation studies on different numbers of extracted knowledge triplets as well as different knowledge encoders. 

\textbf{Numbers of external triplets.}  To analyze the model performance, we conduct experiments to explore the impact of the amount of the knowledge triplets. We evaluate this on visual question answering tasks using OKVQA dataset. Fig. \ref{fig:number} shows the experimental results. Unsurprisingly, increasing the number of retrieved knowledge triplets improve the accuracy of predicted answers. We achive the best accuracy of $52.4\%$ when the number of triplets is 5.

\textbf{Knowledge Encoders.} We also evaluate different knowledge encoders, i.e., different embeddings of implicit or/and explicit knowledge. Table \ref{tab:ablation2} shows the superiority of our model with different encoders. Our model produces 1.8\% and 19.7\% more accurate results than the best and worst performing baselines respectively. Although using ConcepNet embeddings solely, our multimodal training pipeline also learn implicit knowledge in the multimodal fusion encoder. Moreover, we compare our model performance with the models using external knowledge resources. Our proposed model takes advantages of implicit knowledge from large-scale vision-language pretrained models and integrating explicit knowledge prior information. Therefore, we outperform other models using external knowledge resources.


\begin{table}[ht]
\caption{Experimental results on varying the different knowledge encoders on OKVQA dataset. In the literature, these are the baselines that are evaluated on only OKVQA dataset. ``B" denotes base models, ``CN" denotes ConceptNet, and ``GI" denote Google Images. Our model VK-OOD outperforms the baselines on the OKVQA dataset.} \label{tab:ablation2}
\begin{center}
\begin{tabular}{c|c|c}
\hline
\multicolumn{1}{c|}{Method} & \multicolumn{1}{c|}{Knolwedge resources}& \multicolumn{1}{c}{OKVQA} \\ \hline
ConceptBERT & CN & 33.7 \\
KRISP & Wiki + CN & 38.4 \\
MAVEx & Wiki + CN + GI & 39.4 \\
KAT-B & Wiki + GPT3 & 50.6\\
UnifER & CN + ViLT &42.1 \\ \hline
\multirow{2}{*}{VK-OOD} &   CN  &  51.1   \\ 
   &  CN + BERT  &  \textbf{52.4}  \\
\hline
\end{tabular}
\end{center}
\end{table}

\subsection{Robustness of the OOD Layer}
\label{sec:robust_ood}

\textbf{Incomplete Knowledge Triplets.} To evaluate the sensitivity to OOD detection performance, we conduct experiments of incomplete knowledge triplets with missing values. Fig \ref{fig:robustness} shows the results. The green line and blue line are the normalized Hamming distance of $G_{ti}$ (left) and VQA Accuracy (right). Note that, hamming distance of $G_{ti}$ encodes the amount of out-of-distribution knowledge triplets in the data. The green line shows the hamming distance increases with the amount of incompleteness in the data. On the other hand, as expected, the blue line shows that the performance of the model decreases with the amount of incompleteness in the data. However, even with $30\%$ incomplete data, our model achieves higher accuracy ($51.2 \%$) than the best baseline model ($50.6 \%$). Moreover, we measure the impact of the OOD layers on the pretraining objectives. The results are given in the Appendix (see \ref{ap:ablation}).

\textbf{Number of Clusters. } We explore impacts of the number of clusters in optimizing the GMM process and the results are shown in Fig \ref{fig:number}. Note that we present an empirical analysis on the same in Fig. \ref{fig:number} above with the blue line. The general trend is that the performance improve with the number of clusters. To further justify the quality of the learned representation in our model, we illustrate the learned output features by U-MAP \cite{mcinnes2018umap-software} in Fig. \ref{fig:feature}.  Here, we present the feature embedding spaces of multiple modalities on COCO dataset. Different colors indicate different clusters. We show example images of the clusters in Appendix \ref{ap:qa}. Comparing with the baseline, the results demonstrate that more clusters can be identified over the multimodal features extracted by our VK-OOD model. We also show feature maps with different number of clusters in our optimization setups in Fig. \ref{fig:feature}. It implies that our model is able to detect outliers more accurately.





\section{Discussion}
\textbf{Limitations. } In this work, we mainly conduct experiments on discriminative tasks using the encoders only, while generative tasks are still left to be observed. We will extend our model with encoder-decoder architectures to explore the model capabilities on visual question and caption generation tasks. Since we have shown that explicit knowledge can be supervision in vision-and-language training, we believe that different knowledge bases such as medical knowledge graphs are able to provide the user desired domain distributions. We leave these applications and comparisons for future empirical studies. 

\textbf{Conclusions. } We present a training framework to facilitate multimodal analysis under distribution shifts and/or the presence of outlier distributions within the input sample space. There have been various other proposals that either design or exploit special structure in one (or many) of the available modalities for faster pretraining purposes mentioned in \cite {wang2021simvlm}, and for egocentric vision tasks \cite{zeng2022socratic}. While the approaches have been shown to perform well in large-scale settings, from an end-to-end pipeline, those alone may be insufficient -- for example, many frames in a video have low semantic information content and may require complex processing pipelines \cite{cavallaro2005semantic}.  Hence, we believe handling outliers in the context of multimodal analysis is an important topic as more models get integrated or fused. Moreover, none of the models proposed allow for interaction to detect, and (soft) filter them in a computationally efficient way. Naturally, our proposed OOD layer can be directly instantiated within such complex pipelines, while allowing us possibly intervene, and accelerate the training process. We show extensive empirical analysis on various setups asserting that OOD detection in the training pipeline can be extremely effective in downstream tasks. Specifically, we achieve significant training time savings in all our experiments while preserving the state-of-the art performance with respect to many qualitative, and quantitative evaluation metrics.

\medskip

\newpage
\bibliography{bibliography}
\bibliographystyle{abbrv}

\newpage
\appendix

\section{More details of Encoders} \label{ap:encoder}

\subsection{Image Encoder} 

In this work, we focus on patch features and apply vision transformer based models(ViTs) by \cite{vaswani2017attention} as our visual encoder backbones. We split input image into a squence of patches and adopt the linear projection embedding of patch features $V$, which simplifies the step for fusing with text   embedding. We pre-train our model with multiple popular ViTs to examine the influence of image encoder in OOD detection backpropagation process. 

\subsection{Knowledge Encoder}

Given the caption $S$, we parse it into triplets in the form of $T^{id} = \langle o(c),r(c),o'(c)\rangle$, where $o(c)$ and $o'(c)$ are concepts $\in C^{id}$ ,and $r(c)$ is the relation(s) between them, i.e., \textlangle{man, riding, bicycle}\textrangle{}. In our example, the seed triplets(ID triplets) parsed from the caption are $\langle \text{man, riding, bicycle}\rangle$ and $\langle \text{bicycle, down, street}\rangle$. Then we construct knowledge graph by bridging these triplets with external open knowledge including domain and commonsense knowledge graphs, e.g., ConceptNet \cite{speer2017conceptnet}. ConceptNet provides a large scale commonsense knowledge with over 21 million edges by 36 type of relations connecting 8 million nodes, i.e., {IsA, UsedFor, AtLocation}. In this study, to complete our knowledge graph, we collect concepts by querying from ConceptNet using $o(c)$, $o'(c)$ and $rel_i$ where $i \in [0,36]$ and integrate extracted triplets to seed triplets. For example, given ``street" as $o(c)$ and ``AtLocation" as $rel_i$, we will extract the related concepts are located at street to form triple $t_i$. Specifically, we query explicit knowledge triplets of $o(c)$ and $o'(c)$ from ConceptNet to form $T^{cn}$, i.e., \textlangle{bicycle, used for, transport}\textrangle{}. Finally, these knowledge triplets $\in T = T^{cn} \cup T^{id}$ are encoded as language features using a language encoder. 

\subsection{Multimodal Fusion Encoder} 

Considering the above mentioned image $I$ and caption $S$ as inputs, we perform a two-stream transformer pipeline consisting of stacked multiple layers to joint vision and concept text representations. For each layer, we have self-attention unit and merged cross-attention unit which integrates vision and knowledge semantic information and the alignments across them, and a positionwise feed-forward network. 

As the standard transformer architecture in \cite{vaswani2017attention}, the attention function computes identical learnable parameters (weights) as in Eq. \ref{eq:attn1} and Eq. \ref{eq:attn2} , where d is the dimension of the inputs, a query Q, key K, and value V. We use fusion encoder recursively comparing similarity among the image-text pairs as, 
\begin{align}
\label{eq:attn1}
\begin{split}
    \text{Attn}\left(Q_I,K_L,V_L\right) = \\ \text{softmax}\left(\frac{Q_IK_L^T }{\sqrt{d}}\right))V_L,
\end{split}
\end{align} 

and 

\begin{align}
\label{eq:attn2}
\begin{split}
    \text{Attn}\left(Q_L,K_I,V_I\right) = \\ \text{softmax}\left(\frac{Q_LK_I^T }{\sqrt{d}}\right))V_I
\end{split}
\end{align} 

where I and L denote image modality and language modality respectively.

Moreover, we update the image and language embedding outputs of themselves previous layer as queries and concatenate them together as keys and values. To further improve the performance of attention function, we use a multi-head attention which is composed by multiple paralleled attention function in Eq. \ref{eq:attn1} and Eq. \ref{eq:attn2} in each head. The feed-forward layer transform the outputs of multi-head attention through two fully-connected layers with GeLU activation.

\section{Pre-train objectives} \label{ap:pre-train}
We introduce our pre-train objectives in our pipeline in this section, including image text matching (ITM) and masked language modeling (MLM).  

\subsection{Image Text Matching} 

To incorporate both the vision and the language representations, we adopt ITM which is widely used in previous VL studies. Given an image and text of triple pair $\langle v_m,l_n\rangle$, ITM predicts whether they are matched as positive examples or not, and it is a binary classification problem with the loss function in Equation \ref{eq:itm}. We assume that each image and ID triple pair $\langle v_m,l_n\rangle $, as a positive example. The negative pairs are constructed through batch-sampling. 

\subsection{Masked Language Modeling} 

MLM utilizes vision features and text features of ID concepts and relations to predict the masked tokens in the caption sentence $S$. Following most VL models, we randomly masked some tokens in $S$ replacing as $y^{msk}$ and predict them with their visiual and textual features.

\section{More Experiments}
In this section, we show more experiments on ablation studies and qualitative analysis of our proposed VK-OOD models. 

\begin{table}[]
\caption{Ablation study experiment results of VK-OOD model. Baseline denotes vision and knowledge multimodal without explicit knowledge and OOD detection layer. ITM is image-text matching, and MLM is masked language modeling. Results on VQA are on test-dev set. Both downstream results are in zero-shot settings. The bold values mean the best model in the table. Comparing with the baselines, our model with OOD detection layer outperforms on all objectives with two datasets. Training on combinations of objectives improves model performance.}
\centering
\begin{tabular}{l|l|l|ll}
\hline
\multirow{2}{*}{Model} & \multirow{2}{*}{Objectives} & VQA           & \multicolumn{2}{l}{Flickr30k}                       \\ \cline{3-5} 
                       &                             & test-dev      & \multicolumn{1}{c}{TR@1} & \multicolumn{1}{c}{IR@1} \\ \hline
Baseline                    & ITM                         & 70.6          & 82.1                     & 65.6                     \\
Baseline                    & MLM                         & 72.8          & -                        & -                        \\
Baseline                    & ITM+MLM                     & 74.2          & 88.1                     & 74.1                     \\
VK-OOD                & ITM                         & 72.1          & 84.5                     & 69.8                     \\
VK-OOD                & MLM                         & 73.4          & -                        & -                        \\

VK-OOD                & ITM+MLM                     & \textbf{74.8} & \textbf{89.0}            & \textbf{77.2}            \\ \hline
\end{tabular}
\label{tab:abl}
\end{table}

\subsection{Ablation Studies} \label{ap:ablation}
To evaluate our proposed model, we perform more ablations with the default pre-training settings of the baseline and our model mentioned in Section 3.3 of the main paper. We consider different combinations of pre-train objectives in zero-shot settings.

{\bf Pretraining Objectives with OOD Layer.} We observe our model performance on pre-training objectives. Our raw results are presented in Table \ref{tab:abl}. We train on pre-train datasets with $\mathcal{L}_{\text{ITM}}$ in Equation 7, $\mathcal{L}_{\text{MLM}}$ in Equation 8 and $\mathcal{L}$ in Equation 9. The results in Table \ref{tab:abl} show that training on image-text matching and masked language modeling is beneficial for both downstream tasks comparing to the baseline model, especially, there is promising improvements in image retrieval and text retrieval tasks. Thus, it is beneficial to train on both ITM and MLM for filtering outlier concepts and improve performance on downstream tasks. 

{\bf Backbones on downstream tasks.} Since different backbones of image and language encoders may affect model performance, we compare the difference backbone combinations on image and text retrieval tasks on COCO dataset. The results are shown in Table \ref{retrieval}. We observe that CLIP-ViT as vision encoder and RoBERTa as text encoder outperforms other combinations. 

\subsection{Qualitative Analysis} \label{ap:qa}

\begin{figure}
\centering

\subfloat[Cluster 1]{\includegraphics[width=6.5cm]{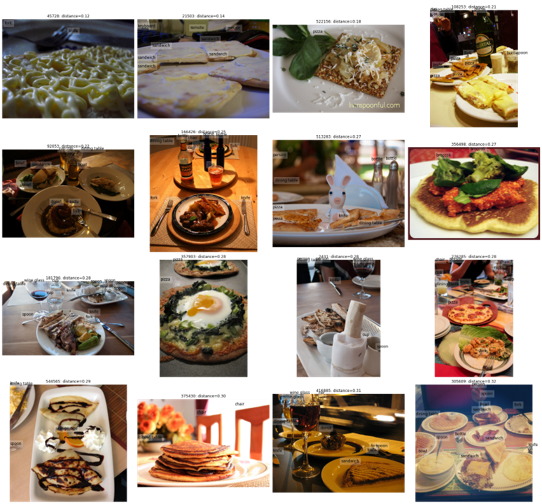}\label{fig:cl1}}
\subfloat[Cluster 2]{\includegraphics[width=6.5cm]{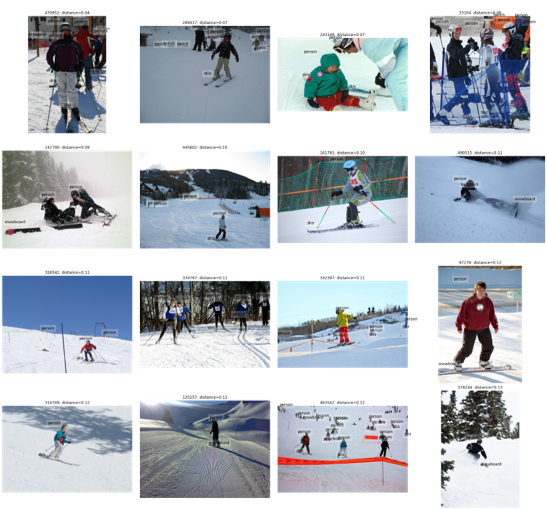}\label{fig:cl2}}

\caption{Example images in each example cluster on COCO val set.}\label{fig:cl}
\end{figure}

Comparing the the baseline, our model result demonstrates more clusters can be identified over the multimodal features extracted by VK-OOD. We also show feature maps with different number of components in our optimization setups in Figure \ref{fig:feature}. Therefore, it illustrates that our model is able to detect outliers and cluster images closest to the corresponding $\mu_i$ with image and explicit knowledge triplets. Figure \ref{fig:cl1} and Figure \ref{fig:cl2} are examples that the nearest images in each cluster.

\begin{figure}
\includegraphics[width=1\textwidth]{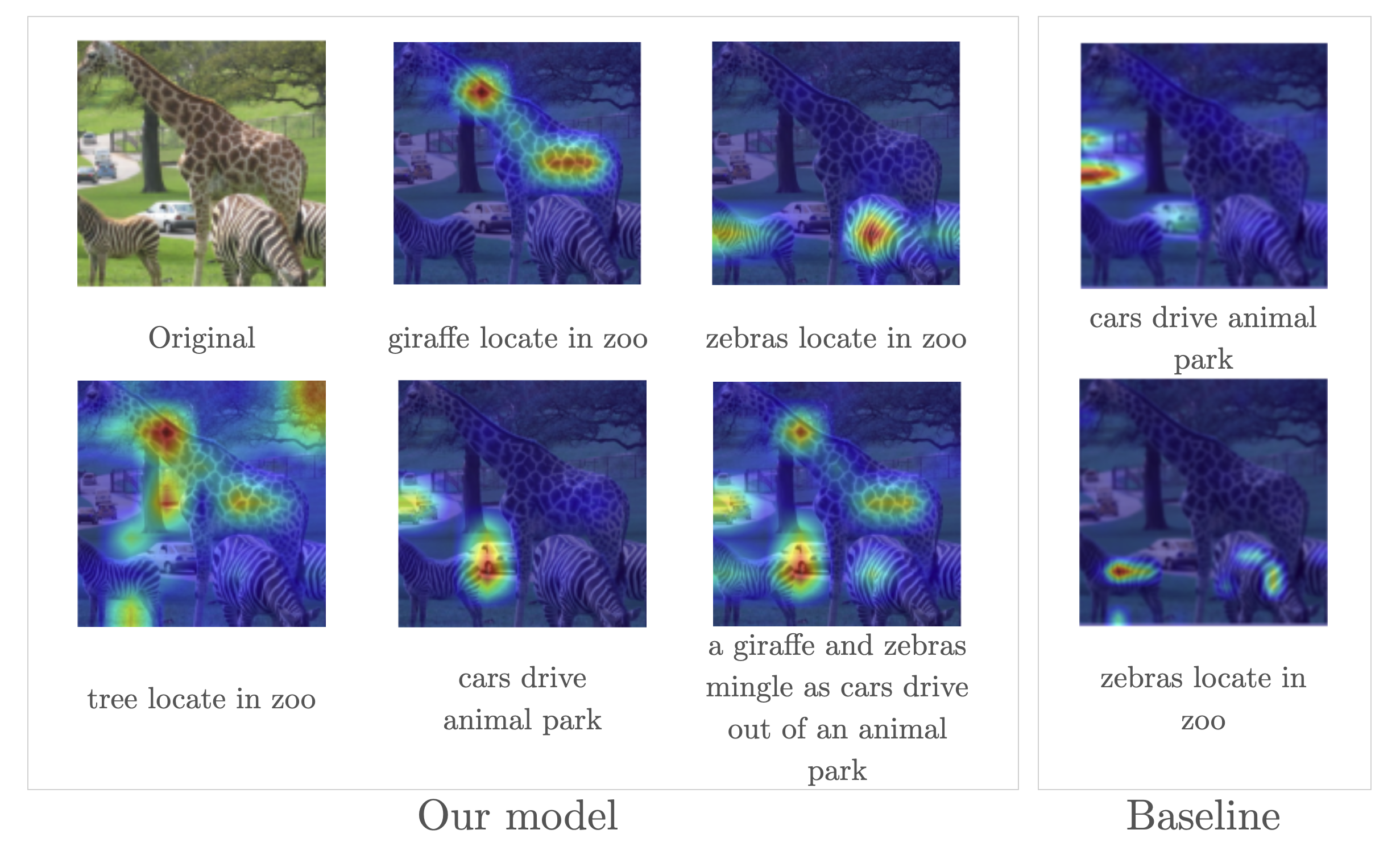}

\centering
\caption{Visualization of the attention maps of image feature $v_m$ and language features $l_{ti}$ knowledge concept triplets alignment. The results are from our VK-OOD model. The original sample caption is ``a giraffe and zebras mingle as cars drive out of an animal park".  We highlight areas in the example image corresponding to different knowledge triplets. Comparing with the attention maps of the baseline model, our model learns object shapes such as zebras and localize those objects correctly. Note that, the baseline model is trained winthout KG and OOD components.\label{fig:example3}%
}
\end{figure}

\begin{figure}
\includegraphics[width=1\textwidth]{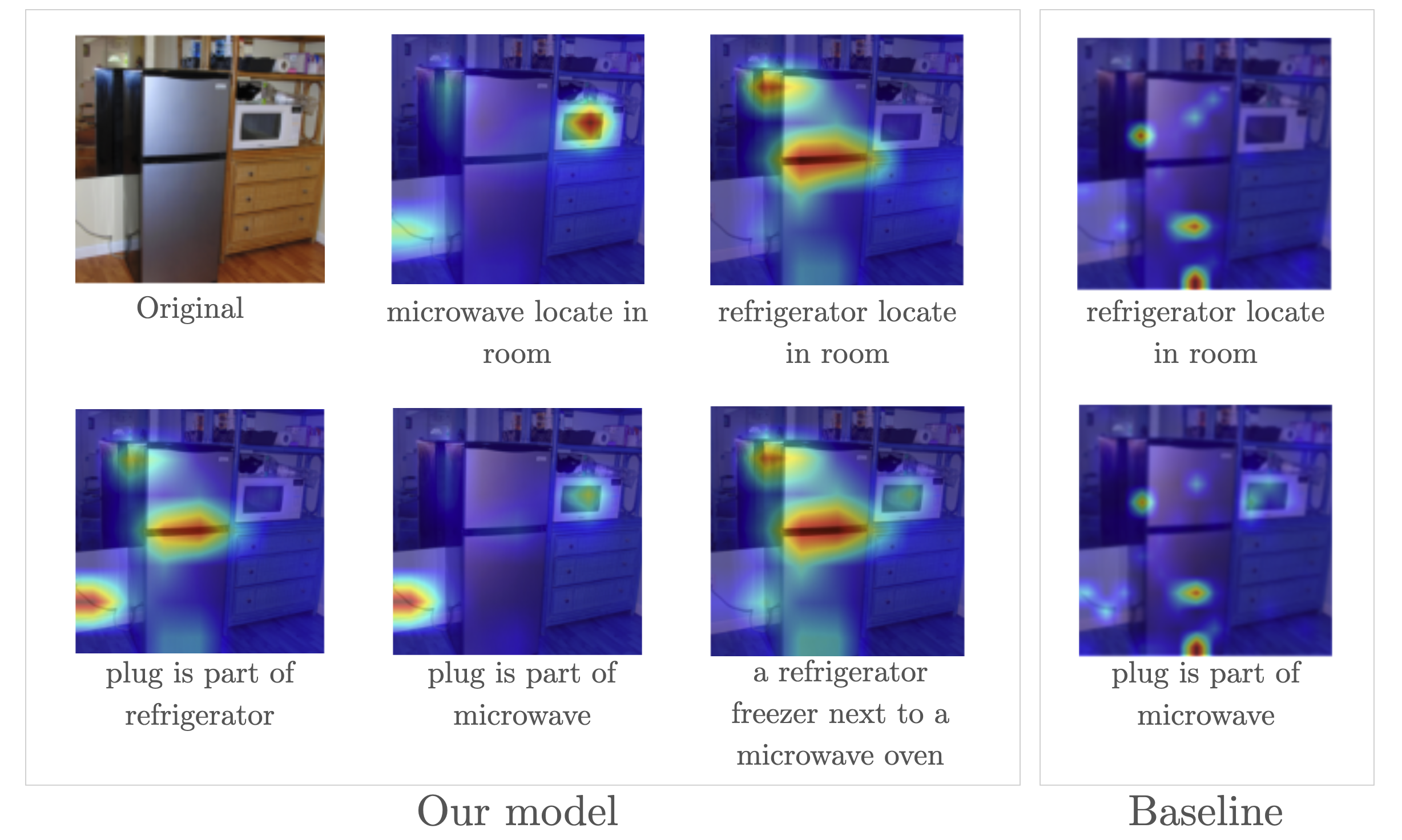}

\centering
\caption{Visualization of the attention maps of image and knowledge concept triplets alignment. The results are from our VK-OOD model. The original sample caption is ``a metallic refrigerator freezer next to a microwave oven". We highlight areas in the example image corresponding to different knowledge triplets. Comparing with the attention maps of the baseline model, our model learns the relations between the parts (i.e., plug) of the objects correctly. Note that, the baseline model is trained winthout KG and OOD components.\label{fig:example4}%
}
\end{figure}

Figure \ref{fig:example3} and Figure \ref{fig:example4} show more qualitative examples of our multimodal alignment results of our pre-train models. We visualize the multimodal attention maps on images corresponding to concept triplets using Grad-cam designed by \cite{selvaraju2017grad}. Following our model architecture, the caption is parsed and integrated with knowledge triplets. The right bottom subfigure in our model in Figure \ref{fig:example3} and Figure \ref{fig:example4} are the multimodal alignment of original captions from MSCOCO \cite{lin2014microsoft} dataset. Other subfigures show the alignments of extracted triplets on the image.

Interestingly, we find that our model is able to capture concept ``plug'' as a part of ``refrigerator'' or ``microwave'' in Figure \ref{fig:example3}. The heatmap area of ``plug'' and ``microwave'' in Figure \ref{fig:example3} clearly suggest that our model has the capability to exploit different relevance between visual and corresponding conceptual text features. By contrast, the baseline results have not shown the relation between plug and microwave. In Figure \ref{fig:example4}, it shows that we detect three zebras comparing with baseline, but counting cars is not performing well as we expected  -- since the size (or scale) of cars is not sufficiently high, and moreover some parts of them are occluded.

\setlength{\textfloatsep}{8pt}

\begin{table}[]
\centering
\caption{Ablation studies on different backbones on image-text retrieval tasks. ``B" denotes base model. We compare different backbones performance on two datasets. Our model with CLIP-ViT and RoBERTa achieves the best results. Comparing with other baseline, all backbones outperform baselines in term of R@1 on COCO dataset.} \label{retrieval}
\begin{tabular}{cc|cccccc}
\hline
\multicolumn{2}{c|}{Backbone}                       & \multicolumn{6}{c}{COCO}                                                                                                                                      \\ \hline
\multirow{2}{*}{Vision} & \multirow{2}{*}{Language} & \multicolumn{3}{c}{Text retrieval}            & \multicolumn{3}{c}{Image retrieval}                     \\
                        &                           & R@1           & R@5           & R@10          & R@1           & R@5           & R@10                   \\ \hline
\multicolumn{2}{c|}{UNITER-B}                       & 64.4          & 87.4          & 93.1          & 50.3          & 78.5          & 87.2                   \\
\multicolumn{2}{c|}{ViLT-B}                         & 61.8          & 86.2          & 92.6          & 41.3          & 72.0          & 82.5                   \\
\multicolumn{2}{c|}{ALBEF(4M)}                      & 73.1          & 91.4          & 96.0          & 56.8          & 81.5          & 89.2                    \\
\multicolumn{2}{c|}{PixelBERT}                      & 63.6          & 87.5          & 93.6          & 50.1          & 77.6          & 86.2                    \\ \hline
Swin                    & RoBERTa                   & 72.1          & \textbf{93.2} & 95.9          & 51.6          & 78.3          & 88.2                    \\
Swin                    & CLIP-ViT                  & 73.8          & 91.4          & 96            & 52.4          & 81.3          & 90.1                    \\
CLIP-ViT                & CLIP-ViT                  & 69.8          & 87.5          & 93.6          & 48.8          & 78.5          & 82.5                    \\
CLIP-ViT                & RoBERTa                   & \textbf{74.7} & 93.1          & \textbf{96.8} & \textbf{57.9} & \textbf{83.6} & \textbf{92.8} \\ \hline
\end{tabular}
\end{table}

\end{document}